\documentclass{article}
\usepackage[utf8]{inputenc}

\title{Brainstorm Generative model}
\author{.}
\date{March 2020}

\usepackage[final]{neurips_2020}

\usepackage{hyperref}       
\usepackage{url}            
\usepackage{booktabs}       
\usepackage{nicefrac}       
\usepackage{amssymb} 
\usepackage{amsmath}
\usepackage{amsthm}
\usepackage{graphicx}
\usepackage{color}
\usepackage{mathtools}
\usepackage{makecell}

\DeclareMathOperator*{\argmin}{arg\,min}

\theoremstyle{definition}

\title{Diffusion models with location-scale noise}

\author{
   Alexia Jolicoeur-Martineau  \\
   Samsung - SAIT AI Lab, Montreal \\
   Canada \\
   \and
   \textbf{Kilian Fatras} \\
   Mila, McGill University \\
   Canada\\
   \AND
   Ke Li \\
   Simon Fraser University \\
   Canada\\
   \and
   \textbf{Tal Kachman} \\
   Radboud University \\
   Netherlands \\
}

\begin{document}

\maketitle

\begin{abstract}
Diffusion Models (DMs) are powerful generative models that add Gaussian noise to the data and learn to remove it.
We wanted to determine which noise distribution (Gaussian or non-Gaussian) led to better generated data in DMs. Since DMs do not work by design with non-Gaussian noise, we built a framework that allows reversing a diffusion process with non-Gaussian location-scale noise. We use that framework to show that the Gaussian distribution performs the best over a wide range of other distributions (Laplace, Uniform, t, Generalized-Gaussian).

\end{abstract}

\section{Introduction}

Diffusion models are powerful generative models that generate high-quality and diverse data. These methods inject Gaussian noise into the data through a Forward Diffusion Process (FDP), and they learn to reverse the process to go from noise to data. There are many ways to define diffusion models: Score-Based Models (SBMs) learn to predict the score (gradient log density), while (non-scored based) diffusion Models (DM) learn to predict the added Gaussian noise in order to remove it from the noisy data. 

SBMs and DMs generally rely on Gaussian noise. A priori, there is no apparent reason why Gaussian noise would be needed as opposed to other types of noise. Very recent works have started exploring non-Gaussian noise. \citet{bansal2022cold} and \citet{anonymous2023iterative} devise their own diffusion-like frameworks to sample from arbitrary distributions by going from dataset 1 to dataset 2; in both papers, they find that using non-Gaussian distributions as the second dataset (instead of Gaussian noise) significantly worsen the quality of the generated data. More related to our work, \citet{deasy2021heavy} shows that SBM, where we learn the score of a Generalized Normal (GN) distribution, leads to significantly worse results when moving away from the Gaussian distribution (which corresponds to the GN distribution with $\beta=2$). In this paper, we aim to answer the question of whether there exist non-Gaussian distributions that perform better than the Gaussian distribution in (non-scored based) DMs. Our work generalizes the DMs with learnable mean and variance by \citet{bao2022estimating, bao2022analytic} to location-scale family noise distributions, and we test this framework on a variety of noise distributions.

\section{Denoising Diffusion Probablistic Models (DDPM)}

Let $x_{0}$ be real data from the data distribution and $z$ be a random sample from a $\mathcal{N}(0,1)$. Assume $t \in [0, 1, \ldots, T]$.

\subsection{Forward process $q(x_{t+1} | x_{t})$}

In Denoising Diffusion Probablistic Models (DDPM) \citep{ho2020denoising}, one define transition steps of the following type: $$ x_{t+1} = \tilde{f}(t)x_{t} + \tilde{g}(t)z,$$
where $\tilde{f}(t)$ is a scaling term for the data and $\tilde{g}(t)$ is a scaling term for the noise.

The noise process is such that $\tilde{f}(t)=\sqrt{\alpha_t}$ and $\tilde{g}(t)=\sqrt{1-\alpha_t}$ for some $\alpha_t \in [0, 1]$. Let $\bar{\alpha_t}=\prod_{s=1}^t \alpha_s$; then by the property of Gaussian distribution, this means that $x_{t} = \sqrt{\bar{\alpha_t}}x_0 + \sqrt{1-\bar{\alpha_t}}z$ and we approximately have that $x_{t} \sim \mathcal{N}(0,1)$. 

At $x_{t}$, we end up with a prior distribution that does not depend on the real data. Since our goal is data generation, we want to reverse the process from noise $x_{t}$ to data $x_{0}$. 

\subsection{Estimation}

We can estimate the joint distribution $q(x_0, x_1, \ldots, x_T)$ with the following parametrization: $$p(x_0, x_1, \ldots, x_T) = p(x_T)\prod_{t=1}^T p_{\theta}(x_{t-1} | x_{t}).$$ It can be shown that optimizing the variational lower bound is equivalent to minimizing $D_{KL}(q(x_t | x_{t-1}) || p_{\theta}(x_{t} | x_{t+1}))$ for all $t \in [1, \ldots, T-1]$. 

From the Markov property, we know that $q(x_t | x_{t-1}) = q(x_t | x_{t-1}, x_0)$. We can use Bayes Rule to obtain a close form for $$q(x_{t-1} | x_t, x_0)=\frac{q(x_t|x_{t-1})q(x_{t-1}|x_0)}{q(x_t|x_0)}$$ given that all three terms of the equation are known and have a close-form.

It can be shown that the variational lower bound optimization can be reduced to minimizing $D_{KL}(q(x_{t-1} | x_t, x_0) || p_{\theta}(x_{t-1} | x_t))$ for all $t \in [2, \ldots, T]$ where $q(x_{t-1} | x_t, x_0)$ is a closed-form Gaussian distribution depending on $x_t$ and $x_0$. The Gaussian distribution has a known variance term that does not need to be estimated. Given that $q(x_{t-1} | x_t, x_0)$ is Gaussian distributed with known variance, its mean is the only parameter left to be estimated.

As shown by \citet{ho2020denoising, nichol2021improved}, directly minimizing the KL divergence works poorly. It can be shown that $q(x_{t-1} | x_t, x_0)$ only depends on $x_0$ and the noise $z$; thus we can instead estimate $\mathbb{E}_{q(x_t|x_0)}[z|x_t]$ and use the close-form solution of $q(x_{t-1} | x_t, x_0)$ to estimate its mean. Thus estimating the expectation of $z$ given $x_t$ is all you need to reverse the diffusion using DDPM.

\section{Generalized denoising diffusion}

\subsection{Forward process $q(x_{t} | x_{0})$}

Contrary to DDPM and other diffusion models, our generalized framework directly samples from $q(x_{t}|x_{0})$ rather than sample from $q(x_{t+1}|x_{t})$ one step at a time. We thus directly assume that $$ x_{t} = f(t)x_0 + g(t)z \sim F(f(t)x_0, g(t)), $$
where $F$ is any distribution of the location-scale (Gaussian, Laplace, Uniform, ...) family, and thus $z$ has a distribution $F(0, 1)$. The noise $z$ corresponds to the added noise/corruption. Similar to most diffusion models, we assume diagonal scaling components, thus i.i.d. noise corruptions.

In our setting, we make no assumptions about the in-between steps $q(x_{t}|x_{t-1})$. In the Gaussian case, the transition steps are just Gaussian. However, when $z$ is non-Gaussian, the distribution of that transition step can be extremely complicated and intractable. Nevertheless, this $q(x_{t}|x_{t-1})$ is unknown and unimportant to us in this framework, as will be seen next.

\subsection{Reverse process}

Our goal is to sample from $q(x_{t-1}|x_{t})$ so that we can reverse the diffusion process from noise to data. However, as mentioned, we do not know $q(x_{t}|x_{t-1})$, so we cannot try to match this term; it also means that we cannot get the close-form solution for $q(x_{t-1} | x_t, x_0)$ using Bayes rule as it depends on the unknown transition probability $q(x_{t}|x_{t-1})$. Thus, we cannot use the original DDPM approach discussed in Section 1. 

We show below how estimating the distribution of the noise $z$ given $x_t$ allows us to directly sample from $q(x_{t-1}|x_{t})$ by plugging the sample from $q(z|x_t)$ into a deterministic equation.

From the forward equation, we know that $$ x_0 = \frac{1}{f(t)} x_t - \frac{g(t)}{f(t)} z. $$
Thus, if we could sample from that $z$ conditional on $x_t$, we could effectively sample from $q(x_0 | x_t)$. 

Furthermore, taking a forward step $q(x_{t-1} | x_0)$ with the same $z$, we get that:
\begin{align}
    x_{t-1} &= f(t-1)x_0 + g(t-1) z \\
    &= \frac{f(t-1)}{f(t)}x_t + \left( g(t-1) - \frac{f(t-1)g(t)}{f(t)} \right) z \\
    &= \bar{f}(t, t-1)x_t + \bar{g}(t, t-1) z,
\end{align}
where $\bar{f}(t, s) = \frac{f(s)}{f(t)}$ and $\bar{g}(t, s) =
g(s) - \frac{f(s)g(t)}{f(t)}$. 

Thus, \emph{by sampling from $q(z|x_t)$, we can deterministically recover a sample from $q(x_{t-1}|x_t)$}. 

\subsection{Estimation}

We can use variational methods to estimate $z$ as $z(x_t)$. Since we know that $z$ is a sample from the distribution $F(0, 1)$ in the forward process, we propose to estimate it as $z(x_t) \sim F(\mu_{\theta}(x_t), \sigma_{\theta}(x_t))$ in the reverse process; this is a generalization of the variational approximation done in Extended-DDPM \citep{bao2022estimating, bao2022analytic}, for the non-gaussian case. Since this is a location-scale family, the reverse steps are approximated as:
\begin{align}
    x_{t-1} &= \bar{f}(t, t-1) x_t + \bar{g}(t, t-1) F(\mu_{\theta}(x_t), \sigma_{\theta}(x_t)) \\
    &= \bar{f}(t, t-1) x_t + \bar{g}(t, t-1)\mu_{\theta}(x_t) + \bar{g}(t, t-1) \sigma_{\theta}(x_t) F(0, 1) \label{eqn:prev}
\end{align}

To estimate the $\mu_{\theta}$ and $\sigma_{\theta}$, the location and scale of the noise distribution, one use KL divergence minimization or equivalently Maximum Likelihood Estimation (MLE). However, similar to \citep{ho2020denoising, nichol2021improved, bao2022estimating, bao2022analytic}, we found this objective generally less numerically stable and impossible to use in some distributions (such as the uniform distribution due to the bounds on the support).

In the non-Gaussian case, KL divergence minimization (or MLE) cannot be solved analytically or lead to complicated equations, making the optimization more challenging and unstable. To solve this issue, we use the Method of Moments (MoM). The MoM seeks to estimate a distribution by matching all the moments $\mathbb{E}[z^k]$, for $k=0, 1, \ldots, \infty$. Thankfully, in the case of location-scale family distributions, we only need two moments $\mathbb{E}[z]$ and $Var[z]$ to estimate the location and scale parameters of the distribution. Thus, all we need is to estimate $\mathbb{E}[z|x_t]$ and $Var[z|x_t]$ and then extract the location and scale terms of the noise distribution. MLE and MoM are equivalent in the Gaussian case, but using the MoM is much more stable and simpler when handling non-Gaussian distributions, so we use it.

Since we can only sample from $q(x_t|x_{0})$, we cannot estimate the expectation from multiple $x_{0}$ given one $x_{t}$ directly. We thus make use of Monte-Carlo by estimating the moments as $$\mathbb{E}[z | x_t] \approx \tilde{\mu}_{\theta_1}(x_t) =  \argmin_{\theta_1} \mathbb{E}_{q(x_t | x_0, z)q(z)}[(z - \tilde{\mu}_{\theta_1}(x_t))^2]$$ and $$Var[z | x_t] \approx \tilde{\sigma}^2_{\theta_2}(x_t) =  \argmin_{\theta_1} \mathbb{E}_{q(x_t | x_0, z)q(z)}[((z - \tilde{\mu}_{\theta_1}(x_t))^2 - \tilde{\sigma}^2_{\theta_2}(x_t))^2].$$ 

From the MoM, for most distributions, we can easily extract the location $\mu_{\theta_1}(x_t)$ and scale $\sigma_{\theta}(x_t)$ parameters from these the approximations of the two moments $\mathbb{E}[z]$ and $Var[z]$. This allows us to easily generalize to most distributions and use the same loss functions in all cases with minimal effort.

As an example, the Laplace distribution has $E[z] = \mu$ and $Var[z] = \sqrt{2}\sigma^2$. Thus $\mu_{\theta_1}(x_t) = \tilde{\mu}_{\theta_1}(x_t)$ and $\sigma_{\theta_2}^2(x_t) = \frac{1}{\sqrt{2}} \tilde{\sigma}^2_{\theta_1}(x_t)$.

\subsection{Similarities and differences to existing DMs}

If we don't estimate $Var[z|x_t]$, our training process for the Gaussian model is equivalent to the one in DDPM \citep{ho2020denoising} and the sampling process to DDIM in which case you estimate the distribution $q(z|x_t)$ using the single value $\mathbb{E}[z|x_t]$. 

When we do estimate a mean and variance, our training process is equivalent for the Gaussian model to the one in Extended-DDPM \citep{bao2022estimating, bao2022analytic}, and our sampling process can be seen as a variational generalization of DDIM since it approximately samples from $q(z|x_t)$. Although both our sampling method and the one in Extended-DDPM can be seen as generalizations of DDIM in the variational case, the generalization of DDIM in Extended-DDPM differs from ours. In Extended-DDPM, they do not sample $q(z|x_t)$ and still use $\mathbb{E}[z|x_t]$ while incorporating $Var[z|x_t]$ separately with additional new noise.

Contrary to other works, our theoretical framework explicitly defines the one-shot forward process $q(x_t|x_0)$, but not $q(x_{t}|x_{t-1})$. We also use the MoM instead of minimizing a KL divergence. Finally, our method generalizes to non-Gaussian distributions using the method of moments.

\section{Results}

We test this framework (GDDIM) on a wide range of location-scale family noise distributions: Gaussian, Student-t, Laplace, Generalized Gaussian ($\beta=1.5$, $\beta=2.5$), and Uniform distributions.

\begin{table}[h]
\centering
\small
\caption{Results on CIFAR-10 with 100 reverse steps.
}
\label{tab:cifar10}
\begin{tabular}{ccccccccc}
\toprule
& & \multicolumn{4}{c}{Heavy Tails} & Medium Tails &  \multicolumn{2}{c}{Light Tails}  \\ Schedule & Sampling & \makecell{t \\ (df = 3)} & \makecell{Laplace \\ $(b = 1)$} & \makecell{\\$(b = 1.5)$} & \makecell{Gaussian \\ $(b = 2)$} & \makecell{\\$(b = 2.5)$} & \makecell{Uniform \\ $(b = \infty)$} \\ \hline
Linear & DDIM & & & & 3.53 & & \\ 
Cosine & DDIM & & & & 5.02 & & \\\hline
Linear & GDDIM & 407.28 & 11.25 & 9.13 & 4.62 & 29.22 & 354.72 \\
Cosine & GDDIM & 340.85 & 10.58 & 14.86 & 4.40  & 26.87 & 274.06 \\
\bottomrule
\end{tabular}
\end{table}

\section{Conclusion}
GDDIM performs similarly, albeit slightly worse than DDIM, but allows non-Gaussian noise distributions. The Gaussian distribution performs better than Non-Gaussian distributions, although the Laplace distribution is a close second. Lighter tails distributions lead to significantly worse performance than heavier tails distributions. Theoretical work is needed to explain the clear advantage of the Gaussian distribution over all other choices of distributions.

\section{Acknowledgment}
We would like to acknowledge Yang Song for his contributions to the paper and for helping shape the ideas and concepts.
T.K would like to acknowledge funding from Lineage Logistics and being hosted by Kells institute. KF was partially supported by the NSERC Discovery grant (RGPIN-2019-06512) and a Samsung grant.
\bibliographystyle{plainnat}
\bibliography{paper}

\appendix

\end{document}